\documentclass[preprint,12pt,numafflabel]{elsarticle}

\usepackage{amssymb}
\usepackage{tabularx}
\usepackage{enumitem}

\usepackage{geometry}
\usepackage{adjustbox}
\usepackage{caption}
\usepackage{amsmath}

\journal{}

\begin{document}

\begin{frontmatter}



\title{Revolutionizing Global Food Security: Empowering Resilience through Integrated AI Foundation Models and Data-Driven Solutions}

\author[inst1]{Mohamed R. Shoaib}

\affiliation[inst1]{organization={School of Computer Science and Engineering, Nanyang Technological University},
            city={Singapore},
            postcode={639798}, 
            country={Singapore}}

\author[inst2]{Heba M. Emara}
\author[inst1]{Jun Zhao}

\affiliation[inst2]{organization={Department of Electronics and Electrical Communications Engineering, Faculty of Electronic Engineering, Menoufia University},
            city={Menouf},
            postcode={32952}, 
            country={Egypt}}

\begin{abstract}
Food security, a global concern, necessitates precise and diverse data-driven solutions to address its multifaceted challenges. This paper explores the integration of AI foundation models across various food security applications, leveraging distinct data types, to overcome the limitations of current deep and machine learning methods. Specifically, we investigate their utilization in crop type mapping, cropland mapping, field delineation and crop yield prediction. By capitalizing on multispectral imagery, meteorological data, soil properties, historical records, and high-resolution satellite imagery, AI foundation models offer a versatile approach. The study demonstrates that AI foundation models enhance food security initiatives by providing accurate predictions, improving resource allocation, and supporting informed decision-making. These models serve as a transformative force in addressing global food security limitations, marking a significant leap toward a sustainable and secure food future.
\end{abstract}




\end{frontmatter}


\section{Introduction}
\label{sec:sample1}

The main goal of ensuring food security is to maintain a sufficient provision of healthy food for global sustenance. This objective was emphasized by the United Nations (UN) in 2015 through the establishment of 17 Sustainable Development Goals (SDGs) to achieve by 2030, aiming to promote global well-being and preserve the Earth's ecosystems \cite{assembly2015resolution}. Among these goals, the second SDG seeks to eradicate hunger by enhancing food security and nutrition and promoting sustainable agricultural practices.

Meeting this objective, however, presents a significant challenge because of the complex nature of food security phenomena. Addressing this complexity necessitates the integration of heterogeneous data spanning different themes, timeframes, and geographical regions. Consequently, the application of techniques that amalgamate these diverse variables becomes imperative to enhance forecasting accuracy.

In recent times, there has been a notable increase in the accessibility of earth observation data, which includes information on various biophysical and climatic aspects. This influx of information, combined with ground-level data such as crop yields, cropland details, crop types, and field delineation, has catalyzed the adoption of Machine Learning (ML) models. These models are employed to automatically distill pertinent features from vast and disparate datasets, enabling high-frequency food security predictions on a global scale \cite{balti2022multidimensional,rhif2022detection}. Leveraging these pertinent features, ML models assume the role of monitoring data and forecasting key aspects of food security \cite{gavahi2021deepyield,qiao2021exploiting,tian2021lstm,luo2022accurately}. In the realm of food security, ML models acquire expertise from diverse and intricate observational data \cite{deleglise2022food,bouras2021cereal}. They meticulously sift through this data, retaining only the salient features that contribute to accurate predictions.

In this context, powerful tools like Long Short-Term Memory networks (LSTM) and Convolutional Neural Networks (CNN) have gained prominence for their application in Deep Learning (DL) methodologies \cite{jong2022improving, zhang2021automated}. These techniques excel at capturing high-dimensional features and appropriately modeling temporal dynamics, resulting in significantly improved prediction accuracy \cite{zhang2020generalized, li2022machine}.

In the rapidly evolving landscape of artificial intelligence and machine learning, a pivotal development in recent years has been the emergence of what is known as a "foundation model." Foundation models, at their core, represent a class of powerful and versatile machine learning models that serve as the bedrock for a wide array of applications across various domains \cite{brown2020language}.

These AI foundation models are typically pre-trained on massive and diverse datasets containing text, images, or both. They leverage deep neural network architectures and extensive computational resources to learn intricate patterns, structures, and semantic relationships present in the input data. The training process involves exposing the model to an extensive range of linguistic and visual contexts, allowing it to develop an understanding of natural language, visual content, and even the nuances of human expression \cite{radford2019language}.

One of the most notable exemplars of foundation models is the Generative Pre-trained Transformer (GPT) series, with GPT-3 being a prominent example. GPT-3, developed by OpenAI, has gained considerable recognition for its impressive capacity to produce coherent and contextually appropriate text, rendering it a versatile instrument for tasks related to natural language understanding and generation. Its proficiency extends from answering questions and language translation to content generation and text completion \cite{gratzer2017mountain}.

The significance of foundation models in the context of food security and related domains lies in their capacity to comprehend and process large volumes of heterogeneous data. By fine-tuning these foundation models on specialized datasets relevant to food security, researchers and organizations can harness their latent capabilities to extract valuable insights, predict food security trends, and offer data-driven policy recommendations.

For instance, when applied to the realm of food security, foundation models can be utilized to analyze vast datasets that include textual reports on crop conditions, satellite images of agricultural regions, weather data, economic indicators, and more. These models excel at distilling actionable information from such data by recognizing patterns and correlations that might elude traditional analysis methods \cite{unicef2021state}.

Furthermore, their natural language understanding capabilities enable them to parse and summarize reports, academic papers, and policy documents, thereby assisting experts in staying updated with the latest research and policy developments in the field of food security \cite{zhang2023umami}.

In summary, AI foundation models represent a transformative tool for enhancing the field of food security. Their innate ability to process, interpret, and generate information from diverse datasets holds the potential to revolutionize the way we address and mitigate food insecurity challenges, ultimately contributing to the achievement of the United Nations' SDGs in the domain of global prosperity and environmental protection. As such, the integration of foundation models into the food security domain exemplifies the convergence of cutting-edge AI technologies with crucial global priorities. Table \ref{tab:TB1} lists the abbreviations or acronyms used in this article.
\begin{table}[]
\centering
\caption{The Abbreviations and/or Acronyms}

\begin{tabular}{p{1.4cm} p{5.8cm}}
\hline
\multicolumn{1}{c}{\textbf{Abbreviations}} & \multicolumn{1}{c}{\textbf{Description}}                                        \\ \hline
UN            & United Nations                      \\
SDGs              & Sustainable Development Goals                        \\
ML                 & Machine Learning                 \\
DL                      & Deep Learning  \\      
CNN    & Convolutional Neural Networks  \\
GPT        &  Generative Pre-trained Transformer  \\
EO            &  Earth observation  \\
RF            & Random Forest \\
SVM           & Support Vector Machine  \\
GDD           & Growing Degree Days  \\
NLP           & Natural Language Processing \\
GRU           & Gated Recurrent Unit \\
LSTM                & Long Short-Term Memory networks    \\
\hline
\end{tabular}
\label{tab:TB1}
\end{table}

\section{A comprehensive overview of machine learning methodologies applied in forecasting tasks related to food security.}
\begin{figure}[htb]
\centering
\includegraphics[width=0.8\textwidth]{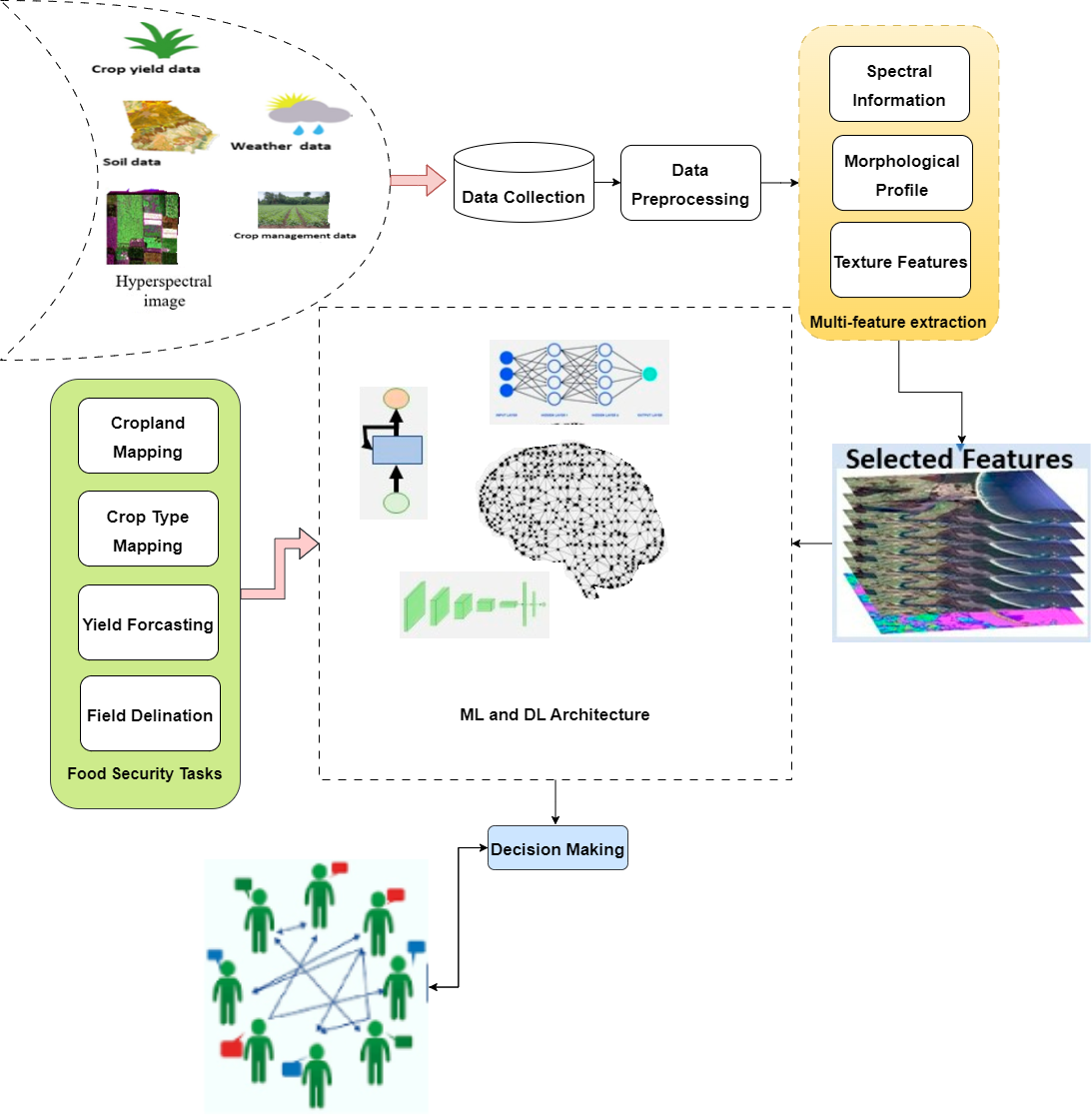} 
\caption{Food Security Process.} 
\label{f1}
\end{figure}

To ensure food security in alignment with the SDGs, a set of four essential sub-tasks is mandated \cite{christopher2021sustainbench}. These tasks involve the mapping of agricultural land \cite{wang2020weakly}, the classification of crop types \cite{m2019semantic}, the prediction of crop yields \cite{you2017deep}, and the delineation of fields \cite{aung2020farm}. The schematic depiction of the food security process is elucidated in Figure \ref{f1}.
National
Despite the rapid evolution of ML techniques over recent decades, it remains imperative to comprehend the procedural intricacies underpinning these methods and rigorously assess their performance. Such scrutiny is vital for enhancing decision-making processes within the domain of food security.

The datasets employed in this study exhibit differences in both spatial and temporal resolutions. To optimize the performance of the machine learning models, the relevant data were meticulously gathered via thorough ground-level observations. These observations serve a dual purpose: validating the resultant model predictions and furnishing essential input variables such as crop yield observations and cropland data.
\subsection{Food Security Tasks}
\subsubsection{Cropland Mapping}
Cropland classification is essential for global food security support, offering valuable insights for monitoring food security \cite{gavahi2021deepyield}. The abundance of earth observation data enables changes in crop mapping through machine learning (ML) classification techniques \cite{abdi2020land,kobayashi2020crop}. These techniques utilize diverse data from various sensors, incorporating parameters like vegetation indices, soil characteristics, moisture content, and climate factors. ML techniques typically leverage these complex datasets along with data obtained from ground-based observations, as demonstrated in previous studies \cite{ji20183d,zhao2019evaluation}.

\subsubsection{Crop Type Mapping}
Crop-type mapping is a critical task for ensuring food security monitoring. Precise and timely prediction of crop types greatly enhances the efficiency of agricultural decision support systems \cite{ghassemi2022designing, massawe2016crop}. ML techniques, which excel in handling heterogeneous and intricate data, are widely employed in this task, operating at regional and global scales \cite{wang2020mapping,pena2019semantic}.

\subsubsection{Crop Yield Prediction}
In order to meet food security objectives and maintain an uninterrupted food supply, it is crucial to make precise predictions of crop yields, particularly on regional and national scales. These predictions can aid decision-makers in devising strategies for imports and support \cite{shahhosseini2021coupling,ruan2022improving}. This task is inherently intricate, influenced by various data types, including soil properties and climate variables. The integration of multispectral and multi-temporal input data, along with ground-observed data, is instrumental in leveraging ML and DL techniques for accurate crop yield prediction \cite{qiao2021exploiting,tian2021lstm}.

\subsubsection{Field Delineation}
The delineation of field borders is essential for improving the accuracy of crop yield forecasting and supporting initiatives for monitoring food security \cite{watkins2019automating,abou2021deep}. Earth observation (EO) data, known for its extensive coverage and high spatiotemporal resolutions, proves particularly effective in this context for field delineation in agriculture. Supervised classification methods are frequently utilized to derive and establish spatial patterns at various scales, including local, regional, and global levels \cite{watkins2019automating}. Moreover, several DL methods have exhibited exceptional proficiency in diverse classification and segmentation tasks, including the demarcation of agricultural fields.

\subsection{Recent Applications of ML and DL in Food Security}
In recent years, the fusion of ML and DL has significantly transformed the food security landscape, profoundly influencing decision-making processes \cite{hossain2019alternatives,biffis2017satellite}. ML techniques have emerged as crucial tools in this area, necessitating substantial amounts of training data. Traditionally, food security assessments heavily relied on surveys conducted by skilled professionals \cite{sood2021computer,razzaq2021automatic}. However, these conventional data collection methods are limited in scope and often involve significant expenses. Earth observation data, climate data, and land use and land cover data are commonly utilized in food security applications. The integration of these diverse data types within ML models has demonstrated promising results. Recently, DL models, including Convolutional Neural Networks (CNN) and Long Short-Term Memory (LSTM) networks, have seen significant use in the field of food security \cite{kuwata2015estimating, kaneko2019deep, wang2020winter}.

 ML techniques encompass three primary learning directions: supervised, unsupervised, and semi-supervised learning. These approaches enable the automatic extraction of features that maximize model performance, allowing the direct training of raw data. However, it is essential to note that these models might be vulnerable to overfitting, especially when dealing with comparatively restricted training datasets.
\begin{enumerate}
    \item {Supervised Learning:}
In supervised learning, the training process depends on labeled data. This data consists of data points, each associated with a specific label or target data. The learning model's goal is to identify the relationship between the input data and the output labels on the unlabeled dataset \cite{cunningham2008supervised}. This relationship allows the model to classify new, unlabeled input data into predefined classes determined during training. In the context of food security applications, the input data might include various types of information like earth observation data, climatic data, and biophysical data, while the output labels may correspond to survey food data and observed ground data, such as cropland and crop type \cite{liu2011web}.

\item{Unsupervised Learning:}
Unsupervised learning takes a different approach, as the training data comprises solely unlabeled data, devoid of specified outputs. Consequently, these techniques are tasked with uncovering hidden structures within the data \cite{dy2004feature}. The training process involves identifying concealed patterns that can categorize the data into subsets with shared characteristics. After identifying and training these subgroups in the training phase, new input data can be classified into implicitly defined categories during the learning process, usually in the testing phase \cite{ben2023unsupervised}.

\item{Semi-supervised Learning:}
becomes relevant in scenarios with scarce labeled training datasets and abundant unlabeled data \cite{van2020survey}. In this scenario, a supervised model is initially trained using the labeled data, subsequently generating additional training data. An unsupervised model is subsequently trained using both the unlabeled data and the labeled data generated by the supervised model. Semi-supervised learning utilizes this combined strategy to address the constraints present in purely supervised or unsupervised methods \cite{yang2022survey}.
\end{enumerate}
\subsection{ML and DL methods for ensuring food security}
Recent advancements in food security have seen the extensive use of both ML and DL techniques. ML methods, such as Support Vector Machines (SVMs), Random Forests (RF), and decision trees, have been widely employed for forecasting tasks in food security. At the same time, DL approaches have demonstrated their efficacy in handling diverse and complex datasets, encompassing both structured and unstructured data.

The utilization of ML methods encompasses multiple steps, including data collection, pre-processing, feature selection, model selection, hyperparameter tuning, training, validation, and testing. This inclusive process streamlines the generation of extensive, varied, and data-rich datasets, providing abundant prospects for expediting research and implementing solutions driven by data \cite{Martini2022}. Various studies in the literature have leveraged ML techniques to monitor food security, with each of the four primary food security tasks demonstrating significant viability when processed using ML methods.

 DL techniques have gained prominence due to their accuracy, especially when dealing with heterogeneous data. However, DL models require extensive computational resources, and their results can be challenging to interpret \cite{Qasrawi2023}. Various CNN architectures, including ResNet and R2U-Net, have played a crucial role in handling tasks related to food security. In conclusion, DL techniques are the preferred choice for effectively analyzing intricate and diverse data, enabling accurate forecasting of the four key food security tasks due to their high precision and ability to handle extensive, unorganized datasets.
\begin{table}[htbp]
    \centering
    \caption{Pros and Cons of ML and DL Techniques}
    \resizebox{\columnwidth}{!}{%
\renewcommand{\arraystretch}{1.8}
    \begin{tabular}{|c|c|c|}
        \hline
        \textbf{Technique} & \textbf{Pros} & \textbf{Cons} \\
        \hline
        LSTM [4, 11] & Flexible, High accuracy, Large dataset processing & Difficult implementation, Sensitive to data amount, Long training time, High computational time \\
        \hline
        XGBoost [46] & Easy implementation, Computational speed, Small data & Fair performance, Assumption dependency \\
        \hline
        RF [45] & Fast Learning, Limited labeled data & Loss, Classic innovation \\
        \hline
        SVM [47] & Simple implementation, Fast learning, Independent of data amount, Good with small-medium dataset & Low computational time, High error, Apt to overfitting \\
        \hline
        CART [61] & Easy implementation, Flexibility, Robust & Low computational time, Structured data \\
        \hline
        K-means [48] & Simple implementation, Training with small data, Flexibility & Structured data, Sensitive to data amount \\
        MLP [80] & Simple implementation, Low computational time & Requires structured data, Apt to overfitting \\
        \hline
        CNN [12, 5, 13] & Simple implementation, Training with small data, Flexibility, Good performance & Complex, Sensitive to data amount, Difficult to interpret results \\
        \hline
        CNN(ResUNet) [10] & Efficient data handling, High performance & Long processing time, Complexity \\
        \hline
        CNN(R2U-Net) [11] & Structured/unstructured data, Efficient for multi-class problems, High accuracy & High computational time, Hard to interpret results \\
        \hline
        KNN [62] & Easy implementation, Robust & Sensitive to kernel parameters, Poor performance on big data \\
        \hline
    \end{tabular}
    \label{tab:ml-pros-cons}
    }
\end{table}
\begin{table}
\centering
\caption{Overview of Research on Food Security Systems (2021)}
\label{comp_2021}
\resizebox{1\textwidth}{!}{%
\renewcommand{\arraystretch}{1.5}
\begin{tabular}{|p{3cm}|p{9cm}|p{9cm}|p{9cm}|p{9cm}|p{9cm}|}
\hline
\textbf{Paper} & \textbf{Methodology} & \textbf{Contributions} & \textbf{Advantages} & \textbf{Disadvantages \& Limitations} & \textbf{Task} \\
\hline
2021 \cite{b1} & Utilizes interviews and surveys to investigate smallholder dairy farms in rural, arid areas of China & Identifies challenges and areas for improvement in smallholder dairy farming in China & Identifies key areas for improvement in smallholder dairy farming, emphasizes the importance of policy support & Indicates slow progress in implementing improvements, particularly in infrastructure and management & Investigates the operations and challenges faced by smallholder dairy farms in China \\
\hline
2021 \cite{b2} & Employs the Fuzzy-TISM methodology to model causal relationships among IoT technologies & Proposes an IoT-driven sustainable food security system for India, focusing on reducing food loss and improving traceability & Offers a holistic approach to sustainable food security, highlights practical implications for policymakers & Acknowledges limitations related to the specific context of India and certain food items & Designs a sustainable food security system leveraging IoT technologies for the Indian context \\
\hline
2021 \cite{b3} & Discusses various AI applications in the food industry with a focus on food processing and quality assurance & Explores the significant role of AI in enhancing food processing and quality assurance, presenting various AI applications in the food industry & Highlights the potential of AI in food security and its various applications across the food industry & Addresses challenges in AI adoption, including cost, integration issues, and proprietary data concerns & Explores the role of AI in enhancing food processing and quality assurance, emphasizing its applications across various aspects of the food industry \\
\hline
2021 \cite{b4} & Systematic review of academic literature on the impact of climate change on food security in Caribbean Small Island Developing States (SIDS) & Identifies challenges faced by Caribbean SIDS due to climate change. Highlights strategies for adaptation. Emphasizes the importance of justice, equity, and local community involvement. & Provides insight into the vulnerabilities of Caribbean SIDS to climate change. Offers a range of adaptation strategies. Emphasizes the importance of community involvement. & Limited body of literature on the topic. Suggests the need for further research. & Analyzes the impact of climate change on food security in Caribbean SIDS and strategies for adaptation. \\
\hline
2021 \cite{b5} & Analysis of social innovation in the food industry in rural China. & Addresses the critical link between food security, poverty alleviation, and social innovation. Showcases examples of social innovation in rural China. & Highlights the role of social innovation in addressing food security and poverty. Discusses various successful initiatives. & Focuses primarily on the rural Chinese context. Suggests the need for continued research in a global context. & Explores the role of social innovation in addressing food security and poverty in rural China. \\
\hline
2021 \cite{b6} & Discussion of the role of genetically modified organisms (GMOs) in addressing food security challenges in Southern Africa. & Emphasizes the significance of food security in Southern Africa. Discusses the potential of GMOs and genome editing in agriculture. Advocates for a comprehensive approach to agricultural productivity. & Offers potential solutions to enhance crop production. Provides an overview of GMO safety and environmental concerns. Discusses the potential of genome editing. & Acknowledges socio-economic and cultural factors hindering GMO adoption. Calls for collaboration, regulatory harmonization, and depoliticization of GM science. & Discusses the role of GMOs in addressing food security challenges in Southern Africa and advocates for responsible use of GM technology. \\
\hline
2021 \cite{b7} & Survey-based study across three campuses, self-reported data, emphasis on nutrition literacy & Highlights the prevalence of food insecurity among college students, emphasizes the impact of nutrition literacy, suggests targeted interventions for diverse student populations & Focus on a specific demographic (college students), suggests practical interventions, aligns with previous research on food insecurity & Reliance on self-reported data and low survey response rates, limited generalizability beyond the college student population & Addressing food security among college students, promoting nutrition literacy, advocating for targeted interventions \\
\hline
2021 \cite{b8} & Online questionnaire study with a focus on Household Dietary Diversity Score and Household Food Insecurity Access Scale, multinomial regression analysis & Examines the impact of the COVID-19 pandemic on food security and dietary diversity, suggests interventions to support vulnerable populations, emphasizes changing consumption patterns & Addresses the impact of a global crisis (COVID-19) on food security, offers actionable recommendations, emphasizes the importance of supporting vulnerable populations & Potential sample bias due to online survey methodology and cross-sectional design, lacks in-depth qualitative analysis & Mitigating the impact of the COVID-19 pandemic on food security, supporting vulnerable populations, recommending policy changes \\
\hline
2021 \cite{b9} & Delphi technique involving 27 expert panel, identification of challenges through interdisciplinary approach & Identifies and categorizes significant obstacles to achieving food security in rural Iran, emphasizes policy, economic, and infrastructural challenges, advocates for a comprehensive approach & Utilizes an expert panel for comprehensive analysis, emphasizes the need for multi-sectoral approaches, provides specific suggestions for policy and program improvements & Relatively small expert panel, possible subjectivity in the Delphi technique, limited generalizability beyond the rural context & Identifying and addressing challenges to food security in rural Iran, advocating for a multi-sectoral approach, suggesting policy improvements and technological advancements \\
\hline
2021 \cite{b10} & Literature review, emphasizing the lack of evidence-based programs & Highlighting the prevalence of food insecurity among college students, emphasizing the need for research and intervention-specific programs & Raises awareness of an understudied issue, calls for collaboration and intervention-specific research & Does not provide specific intervention strategies, may lack primary data analysis & Addressing food insecurity among college students \\
\hline
2021 \cite{b11} & Data analysis using statistical methods like Foster–Greer–Thorbecke measure and probit regression & Identifying factors influencing food security among smallholder maize farmers in Nigeria, suggesting policy recommendations & Provides valuable insights for policymakers, suggests specific policy recommendations based on findings & Limited in scope to smallholder maize farmers, may not fully capture broader agricultural issues & Analyzing factors influencing food security among smallholder maize farmers in Nigeria \\
\hline
2021 \cite{b12} & Data envelopment analysis (DEA), tobit regression, and logit regression & Revealing the low technical efficiency of rice farming in East Java, emphasizing the impact on food security, and proposing strategies for improvement & Provides an in-depth understanding of technical inefficiencies, highlights the importance of government support and involvement of educated individuals in farming & Relatively narrow focus on rice farming in East Java, may not be fully generalizable to other regions & Examining technical efficiency of rice farming and its impact on food security in East Java \\
\hline
2021 \cite{b13} & Binary probit models, recursive framework & Analyzes coping strategies and their relation to food security & Considers interactions between multiple shocks, emphasizes need for support & Does not delve into the detailed effects of other coping strategies like labor deployment or consumption alteration. Limited to the study context. & Examines the impact of shocks on coping strategies and food security in developing countries. Suggests policy measures to support rural farm households. \\
\hline
2021 \cite{b14} & Focus groups, interviews, national content experts & Examines food insecurity among AI/AN adults with T2D & Provides insights into complex food insecurity issues, highlights need for tailored interventions & Findings may not be generalizable beyond the studied sample. Participants had access to comprehensive diabetes education and care programs, which may not represent all AI/AN populations. & Explores food insecurity and its impact on healthy eating in AI/AN adults with T2D. Advocates for location-specific and culturally sensitive interventions. \\
\hline
2021 \cite{b15} & Panel data analysis, system Generalised Method of Moments (GMM) & Explores the impact of social inclusion and innovation on food security in West Africa & Offers insights into the role of social inclusion, innovation, and government effectiveness in food security & Acknowledges data constraints. Calls for more research at the country and household levels for a better understanding of food security dynamics in West Africa. & Investigates the relationship between social inclusion, innovation, and food security in West Africa. Suggests policies to enhance food security and meet SDGs. \\
\hline
2021 \cite{b16} & Panel data analysis using econometric models including POLS, Fixed Effect, and GMM & Highlights the influence of governance and ICT adoption on food security in West Africa, suggests both factors are crucial for achieving food security & Provides robust results using GMM, aligns with SDG-2, underscores positive interaction between governance and ICT adoption for food security & Acknowledges the need for further research at the country or household level, consideration of other dimensions of food security, and the use of more accurate ICT indicators & Analyzing the impact of governance and ICT adoption on food security in West Africa \\
\hline
2021 \cite{b17} & Simple random sampling, structured questionnaires, and key informant interviews, endogenous switching probit model & Emphasizes the significance of off-farm activities in enhancing household food security in rural Ethiopia, suggests policies focusing on on-farm and off-farm sectors & Offers insights into the impact of various socio-economic factors on off-farm participation and food security, recommends specific policy measures & Limited generalizability, potential biases in self-reported data, reliance on a specific model, possible response biases & Investigating the influence of off-farm activities on food security status in rural Ethiopia \\
\hline
2021 \cite{b18} & Stratified sampling method, validated questionnaire, discriminant analysis, identification of five key components of adaptation capability & Investigates the connection between climate change adaptation and household food security in western Iran, emphasizes the need for accurate planning and adaptation in rural areas & Identifies five key components influencing food security, emphasizes the role of technology and rural development, highlights the significance of climate change adaptation & Acknowledges limitations related to sample selection and data sources, underscores the need for future research in the area & Examining the crucial connection between climate change adaptation and household food security in western Iran \\
\hline
2021 \cite{b19} & Introduces WEF metrics (EF, EBC, eco-balance), WEFEB nexus index & Introduces a new WEFEB nexus index, highlights ecological imbalances, emphasizes policy recommendations & Integrated perspective in WEF management, specific policy recommendations & Limited to well-developed African countries, may require further validation of metrics & Investigates ecological sustainability in well-developed African countries \\
\hline
\end{tabular}
}
\end{table}
\begin{table}
\centering
\caption{Overview of Research on Food Security Systems (2022)}
\label{comp_2022}
\resizebox{1\textwidth}{!}{%
\renewcommand{\arraystretch}{2}
\begin{tabular}{|p{3cm}|p{9cm}|p{9cm}|p{9cm}|p{9cm}|p{9cm}|}
\hline
\textbf{Paper} & \textbf{Methodology} & \textbf{Contributions} & \textbf{Advantages} & \textbf{Disadvantages \& Limitations} & \textbf{Task} \\
\hline
2022 \cite{b20} & ESAI, random forest model, analysis of land use changes & Predicts potential land degradation risk, emphasizes monitoring and socio-economic factors & Prediction of future degradation trends, consideration of socio-economic factors & Data availability and spatial resolution constraints, need for further research & Evaluates land degradation in the North China Plain and its implications for food security \\
\hline
2022 \cite{b21} & Comprehensive review of 30 studies, assessment of food insecurity prevalence & Identifies prevalence of food insecurity among AI/AN populations, emphasizes cultural sensitivity, suggests areas for future research & Highlights the need for culturally relevant research, community-based interventions & Reliance on existing studies, challenges in developing culturally relevant survey instruments & Conducts a comprehensive review of food insecurity among AI/AN communities \\
\hline
2022 \cite{b22} & Emphasizes the use of advanced AI technologies, particularly machine learning. & Highlights the potential of AI in predicting and mitigating climate change impact on food safety. & Efficient data analysis and prediction, automation of food quality assessment. & Social, ethical, and legal concerns related to AI usage. Need for data collection initiatives and responsible implementation. & Addressing challenges posed by climate change on food safety. \\
\hline
2022 \cite{b23} & Relies on comprehensive frameworks and observation networks, along with remote sensing techniques and spatial modeling. & Offers an integrated conceptual framework for understanding the CDF nexus in Africa and emphasizes sustainable practices. & Comprehensive understanding of the interplay between climate change, drylands, and food security. & Challenges and uncertainties in climate models for aridity predictions. & Understanding the climate–drylands–food security (CDF) nexus in Africa. \\
\hline
2022 \cite{b24} & Utilizes data-driven Random Forest and Chi-Square Automatic Interaction Detection (CHAID) decision tree methodology. & Provides insights into the complex issues of food security in West-African rural communities and suggests tailored interventions. & Accurate classification of food security levels, identification of key factors, and implications for tailored interventions. & Slight variations in misclassification rates in machine learning models, need for further research into decision tree classifiers. & Identifying factors influencing food security in West-African rural communities and suggesting tailored interventions. \\
\hline
2022 \cite{b25} & Econometric techniques, ARDL model, diagnostic tests & Confirms Green Revolution impact on food security in Pakistan & Robust statistical analysis, policy implications for agricultural productivity improvement & Relies heavily on econometric analysis, may not account for all variables & Understanding the impact of Green Revolution on food security \\
\hline
2022 \cite{b26} & Field experiment in 2021 & DSR improves rice productivity, resilience to extreme rainfall & Enhanced rice productivity, resilience to adverse climatic conditions & Limited time frame of field experiment, specific to rice production & Proposing DSR as a sustainable approach for rice production \\
\hline
2022 \cite{b27} & Review and analysis of existing literature, emphasis on biofertilizers & Biofertilizers as sustainable and eco-friendly alternative to chemical inputs & Sustainable agriculture, increased crop yields, improved nutrient content & Limited discussion on challenges in large-scale implementation, varying effectiveness for different crops & Advocating for the use of biofertilizers for sustainable agriculture \\
\hline
2022 \cite{b28} & Longitudinal survey of 167 American Indian adults & Urgent need for policies and programs for food assistance in American Indian communities. & Highlights gender disparities and the impact of household size on food insecurity. & Limited to a specific community, may not be fully generalizable. & Investigating the trajectories of food insecurity in relation to COVID-19 incidence within a specific community. \\
\hline
2022 \cite{b29} & Multistage sampling technique and structured interviews with 204 sample farmers & Understanding the factors influencing AI technology adoption in dairy farming. & Identifies factors influencing the adoption of AI technology and provides recommendations for improving adoption rates. & Limited to a specific geographical region, may not be fully applicable in other contexts. & Understanding the adoption and intensity of adoption of AI technology in a particular region for dairy farming. \\
\hline
2022 \cite{b30} & Discussion and analysis of the role of Industry 5.0 technologies & Highlighting the role of ICTs, particularly Industry 5.0 technologies, in improving food security and quality in the supply chain. & Provides a comprehensive overview of various Industry 5.0 technologies and their potential applications in the food supply chain. & Primarily discusses the potential of emerging technologies, may not provide specific implementation strategies. & Discussing the potential applications of Industry 5.0 technologies for enhancing food security, quality, and sustainability in the food supply chain. \\
\hline
\end{tabular}
}
\end{table}
\begin{table}
\centering
\caption{Overview of Research on Food Security Systems (2023)}
\label{comp_2023}
\resizebox{1\textwidth}{!}{%
\renewcommand{\arraystretch}{2}
\begin{tabular}{|p{3cm}|p{9cm}|p{9cm}|p{9cm}|p{9cm}|p{9cm}|}
\hline
\textbf{Paper} & \textbf{Methodology} & \textbf{Contributions} & \textbf{Advantages} & \textbf{Disadvantages \& Limitations} & \textbf{Task} \\
\hline
2023 \cite{b31} & Integrating satellite imagery and machine learning techniques & Developing a tool for assessing crop health and predicting food security in Sub-Saharan Africa & Offers a scalable solution for food security assessment using advanced technology, potential for timely interventions & Limited availability of high-quality satellite imagery, challenges related to training machine learning models & Creating a tool for assessing crop health and food security in Sub-Saharan Africa \\
\hline
2023 \cite{b32} & Conducting surveys and using statistical analysis & Investigating the impacts of market integration and land tenure security on household food security in rural Ethiopia & Provides empirical evidence on the relationship between market integration, land tenure security, and food security & Data limitations and potential endogeneity issues, focus on a specific context & Analyzing the effects of market integration and land tenure security on food security in rural Ethiopia \\
\hline
2023 \cite{b33} & Combines machine learning and remote sensing techniques & Predicting crop yields and identifying factors affecting food security in Southeast Asia & Offers a data-driven approach to food security assessment, timely yield predictions & Challenges in data acquisition and model accuracy, reliance on historical data & Predicting crop yields and understanding factors affecting food security in Southeast Asia \\
\hline
2023 \cite{b34} & Employing econometric modeling and surveys & Analyzing the impact of climate change on food security in Central America & Contributes to the understanding of climate change effects on food security, policy implications & Limited to a specific geographical area, challenges in modeling complex climate interactions & Investigating the impact of climate change on food security in Central America \\
\hline
2023 \cite{b35} & Review and analysis of existing literature & Examining the role of blockchain technology in enhancing food traceability and security & Provides an overview of the potential of blockchain technology for food traceability, safety, and supply chain transparency & Primarily discusses the potential of blockchain, limited to the technology aspect & Discussing the role of blockchain technology in enhancing food traceability and security \\
\hline
2023 \cite{b36} & Field surveys and data analysis & Investigating food security and dietary diversity in urban slums of a developing country & Contributes to understanding urban food security challenges, emphasizes the role of dietary diversity & Limited to a specific urban context, may not fully represent rural or global food security issues & Analyzing food security and dietary diversity in urban slums of a developing country \\
\hline
2023 \cite{b37} & Data analysis and statistical modeling & Assessing the impact of agricultural extension services on food security in a specific region & Provides empirical evidence on the effects of extension services, policy implications & Limited to a specific region, challenges in attributing causality & Investigating the impact of agricultural extension services on food security in a specific region \\
\hline
2023 \cite{b38} & Review of existing literature and data analysis & Examining the potential of insect farming in improving food security and sustainability & Provides insights into the role of insect farming, especially in protein production & Limited to specific aspects of food security related to insect farming, potential cultural and regulatory challenges & Discussing the potential of insect farming in improving food security and sustainability \\
\hline
2023 \cite{b39} & Combining econometric modeling and surveys & Analyzing the factors influencing food security in conflict-affected regions & Provides empirical evidence on the relationship between conflict, displacement, and food security & Challenges in data collection in conflict-affected areas, focus on specific contexts & Investigating the impact of conflict and displacement on food security in conflict-affected regions \\
\hline
2023 \cite{b40} & Integrating geospatial data and modeling & Assessing the vulnerability of food supply chains to climate risks & Offers a tool for understanding the climate vulnerability of food supply chains, supports resilience planning & Reliance on geospatial data availability, model uncertainties & Evaluating the vulnerability of food supply chains to climate risks \\
\hline
\end{tabular}
}
\end{table}
\section{ Datasets Utilized for Food Security Tasks}
The area of food security relies on a range of datasets, such as earth observation data, climatic data, biophysical data, and observed ground data. These datasets form the basis for utilizing machine learning (ML) techniques in activities related to food security.

In the realm of ML, various techniques have been developed, each possessing distinct strengths and weaknesses \cite{Mersha2022}. Random Forest (RF) is known for its swift learning, ease of implementation, and computational efficiency, making it particularly advantageous when dealing with limited labeled data. However, its conservative approach to loss optimization and potential for subpar performance due to underlying assumptions should be considered. Meanwhile, XGBoost offers simplicity in implementation, computational speed, and adaptability to smaller datasets, though it may incur longer computation times and be susceptible to higher error rates and overfitting. Support Vector Machine (SVM), with its ease of use, rapid learning, dataset size independence, and proficiency with moderate-sized datasets, provides a compelling option, yet it may involve lengthier computations, a risk of elevated error rates, and a propensity for overfitting. K-means, though straightforward to implement, exhibit sensitivity to kernel function parameters and are less suited for extensive datasets \cite{Vítor2023}. Finally, CNN is praised for its simplicity, flexibility, strong performance, and automated feature selection. Nevertheless, they may encounter challenges with structured data, demonstrate sensitivity to data volume, and yield results that pose interpretation difficulties. Selecting the most appropriate ML technique for a given task necessitates careful consideration of the dataset's characteristics and the project's objectives and constraints.

Food security datasets often contain diverse information, including thematic, structural, and spatio-temporal data. Thematic data encompasses aspects such as Earth, climate, economic, and soil characteristics, while structural data can consist of survey values, vectors, time series, and rasters \cite{Krishnan2023}. Spatio-temporal information varies in terms of spatial resolutions (e.g., region, commune, country, or pixel) and temporal resolutions (e.g., year, month, or week). Integration of these datasets typically requires harmonization techniques to ensure uniformity across different scales. Previous studies have employed various data processing methods to forecast food security at different temporal or spatial levels. Table \ref{t2} provides a comprehensive breakdown of essential data categories and variables for monitoring and assessing food security. 

It is organized into four main categories: "Remote Sensing Data," which includes data on vegetation health, temperature, and precipitation; "Climate Data," encompassing information on rainfall patterns, temperature records, and climate anomalies; "Biophysical Data," consisting of data related to crop types, health, soil properties, and yields; and "Ground-Observed Data," which covers on-the-ground surveys, market prices, food consumption, livestock information, and water resources \cite{Emara2022}. Each variable within these categories serves as a critical component in evaluating and understanding the factors affecting food security in a region. This structured approach to data collection and analysis is instrumental in supporting decision-making processes aimed at improving food access, availability, and overall food security for vulnerable populations \cite{Zeb2022}.

\begin{table}[htbp]
    \centering
    \caption{Input Variables for Food Security Monitoring}
    \label{t2}
    \small
     \resizebox{.8\textwidth}{!}{%
\renewcommand{\arraystretch}{1}
    \begin{tabular}{|p{2.5cm}|p{6cm}|p{6cm}|}
        \hline
        \textbf{Category} & \textbf{Variables} & \textbf{Description} \\
        \hline
        \multicolumn{3}{|c|}{\textbf{Remote Sensing Data}} \\
        \hline
        NDVI & Normalized Difference Vegetation Index & Measures vegetation health and crop growth. \\
        EVI & Enhanced Vegetation Index & Provides improved sensitivity in dense vegetation areas. \\
        LST & Land Surface Temperature & Monitor temperature variations affecting crops. \\
        NDWI & Normalized Difference Water Index & Identifies water bodies and soil moisture levels. \\
        Land Cover & Land Use Classification & Classifies land types (cropland, forests, etc.). \\
        Rainfall & Rainfall Estimates & Monitors precipitation patterns. \\
        Soil Moisture & Soil Moisture Content & Provide soil moisture levels for agriculture. \\
        \hline
        \multicolumn{3}{|c|}{\textbf{Climate Data}} \\
        \hline
        Rainfall & Rainfall Patterns & Historical and current rainfall data. \\
        Temperature & Temperature Data & Historical and real-time temperature records. \\
        Evapotranspiration & Evapotranspiration Data & Measures water loss from soil and plants. \\
        Climate Anomalies & Abnormal Climate Patterns & detect climate anomalies affecting crops. \\
        Growing Degree Days & Growing Degree Days (GDD) & Predicts crop development stages. \\
        Climate Indices & Climate Indices (e.g., PDSI, SPI) & Assess drought severity and precipitation patterns. \\
        \hline
        \multicolumn{3}{|c|}{\textbf{Biophysical Data}} \\
        \hline
        Crop Type & Crop Type and Distribution & Information on types of crops planted. \\
        Crop Health & Crop Health Data & Data on crop diseases, pests, and overall condition. \\
        Soil & Soil Data & Properties such as texture, pH, and nutrient levels. \\
        Crop Yield & Crop Yield Data & Historical and real-time crop yield information. \\
        Vegetation Cover & Vegetation Cover & Percentage of land covered by vegetation. \\
        \hline
        \multicolumn{3}{|c|}{\textbf{Ground-Observed Data}} \\
        \hline
        Crop Surveys & Crop Surveys & Assess crop health, growth, and yield estimates. \\
        Market Prices & Market Prices & Local prices for food items to gauge affordability. \\
        Food Consumption & Food Consumption Surveys & Data on household food consumption and dietary diversity. \\
        Livestock Data & Livestock Data & Information on livestock health and conditions. \\
        Water Resources & Water Resources & Availability and quality of water sources for agriculture. \\
        \hline
    \end{tabular}
    }
\end{table}

\section{Challenges and Prospects for the Future}
Although ML techniques have demonstrated promising outcomes in tackling food security issues, it is essential to address several challenges and future considerations in this regard \cite{Nouredine2023}.

\subsection{Data Challenges}
\subsubsection{Data Complexities}
Food security data come in various formats, such as earth observations, climate data, biophysical changes, and ground-level information. Prior to employing ML techniques, data must undergo collection, cleaning, and validation \cite{Martini2022}. The heterogeneous nature of these data, along with limitations in data processing methods, presents challenges. Integrating data from various sources and structures, such as yield production, crop types, and cropland mapping, remains a challenge.

\subsubsection{Sample Size}
A sufficient number of samples are required to ensure generalizability and avoid overfitting. In food security applications, establishing an adequate training set, including yield production samples or cropland maps, is vital \cite{Martini2022}. The sample size depends on the study's scope, and data collection's diversity and heterogeneity add to the challenge. Choosing an initial sample size that is sufficiently large to avoid overfitting and improve accuracy remains an ongoing challenge.

\subsubsection{Data Availability}
ML approaches are heavily reliant on the accessibility and accuracy of data. Nevertheless, numerous studies concerning food security are constrained by small datasets, frequently confined to local or regional contexts, resulting in challenges such as absent or deficient records. This hampers accurate forecasting \cite{Nouredine2023}. Food security data, including cropland mapping and yield estimation, often suffer from incomplete records. Expanding datasets globally and addressing missing data challenges through methods like data imputation is necessary.

\subsection{Analyzing the Challenges in Computation}
\subsubsection{Selecting the Optimal Machine Learning Model}
Determining the most appropriate ML model frequently requires a thorough assessment of the distinctive attributes of current models \cite{Nouredine2023}. Simplicity and ease of use are preferred, especially when computational resources are limited. Models like Decision Trees and Random Forests are favored for their efficiency, ability to provide valuable feedback, and avoidance of overly complex designs. The choice of model should align with available computing resources. Model interpretability also holds a crucial position in the selection process, where deep learning (DL) models, such as neural networks, are specifically employed for tasks demanding comprehensive scrutiny of extensive datasets, such as remote sensing data captured at frequent intervals and daily climatic station data.

\subsubsection{Feature Extraction and Selection}
Feature selection and extraction represent crucial stages in the implementation of ML algorithms. The availability of features is imperative for accurate forecasting \cite{Nouredine2023}. Integration of feature selection with resampling techniques is common practice, especially when dealing with training datasets. Notably, certain features, such as vegetation and climate data, contain vital information relevant to applications in food security. The selection of specific data features markedly influences the outcomes of the forecasting process.

\subsubsection{Hyperparameters Selection}
ML techniques are dependent on hyperparameters that require optimization for individual datasets to maximize precision \cite{Nouredine2023}. End-users fine-tune these parameters to improve the predictive capabilities of the ML algorithm. Employing tools such as grid search facilitates the optimization of hyperparameters, while the integration of nested resampling methods is vital in preventing overfitting within the ML model.

\subsubsection{Model Complexity}
ML techniques face challenges with big data due to computational complexity. As data scale increases, basic operations consume significant time and memory resources \cite{Nouredine2023}. For instance, traditional ML techniques become computationally infeasible with large datasets, such as fine spatial resolution cropland mapping. The computational time required for these models grows exponentially with data size. Deep learning models prove more efficient than traditional ML as data volume expands.

\subsubsection{Cutting-Edge Machine Learning Approaches}
To tackle the obstacles associated with data heterogeneity and limited sample size, it is imperative to delve into the latest developments in ML techniques. Multi-task learning, transfer learning, and active learning are strategies that can enhance ML methods \cite{Nouredine2023}. These approaches promote knowledge sharing between tasks, leveraging unlabeled data, and involving experts in model building. They hold promise for improving forecasting accuracy in food security applications.

\subsection{Difficulties in Interpretability and Assessment}

\subsubsection{Evaluation Metrics and Uncertainty}
ML techniques leverage diverse parameters to minimize disparities between model predictions and training data. Assessing the effectiveness of complex models necessitates the careful consideration of evaluation metrics. The selection of suitable evaluation metrics is contingent upon the nature of the data, the problem statement, and the specific ML models employed. Given the inherent uncertainty in training data and labels, the application of statistical techniques becomes imperative in evaluating performance discrepancies under uncertainty \cite{Nouredine2023}. Commonly used evaluation metrics in food security ML applications include R, R², RMSE, MAE, and BIAS, each emphasizing different aspects of model performance.

\subsubsection{Reproducibility and Replicability}
Ensuring reproducibility and replicability in data-driven food security research has garnered significant attention. Challenges arise when applying DL technology and large datasets across diverse geographic locations for tasks like yield production and cropland delineation. Ensuring consistency in datasets, methods, and workflows is essential for addressing these challenges \cite{Nouredine2023}.
\section{Large AI Models}
 Various types of large AI models are revolutionizing the approach to food security challenges worldwide. These expansive AI systems, ranging from CNNs to RNNs and transformers, are finding application across the entire food supply chain. In agriculture, they're harnessed to forecast crop yields, analyze climate patterns, and aid farmers in making data-driven decisions \cite{Brown2020} \cite{Vaswani2017}. Reinforcement learning models are optimizing resource allocation, and ensuring judicious use of water and fertilizers, thereby promoting sustainable and efficient farming practices. Meanwhile, generative adversarial networks and natural language processing models enhance food distribution and logistics, curbing spoilage and waste. The diverse capabilities of these large AI models collectively advance food security, facilitating better access to nourishing and dependable food sources for a larger global population.

Foundation models, a pivotal class of large-scale AI models, represent a core element in the realm of NLP \cite{Vaswani2017}. These models are characterized by their pre-training on extensive text corpora, which equips them with an understanding of the intricate dynamics of human language, context, and semantics. At their core, foundation models, exemplified by GPT-3 and BERT, rely on transformer architectures, featuring multi-layered encoders that enable them to capture intricate word, phrase, and sentence relationships \cite{Brown2020}. The pre-trained models can subsequently undergo fine-tuning for specific NLP tasks, making them adaptable for tasks like machine translation, sentiment analysis, and question-answering. The mathematical foundation of these models often denoted as ELMo, GPT, or BERT, has revolutionized NLP, effectively enhancing natural language understanding and generation through their comprehensive representation learning capabilities.

The central mathematical component of foundation models is the transformer architecture, which includes self-attention mechanisms. The concept of self-attention can be described as illustrated in Zhao et al. (2023) \cite{zhao2023convolutional}:

\begin{equation}
\text{Attention}(Q, K, V) = \text{softmax}\left(\frac{QK^T}{\sqrt{d_k}}\right)V
\end{equation}

Where $Q$, $K$, and $V$ are the query matrix, the key matrix, and the value matrix, respectively, and $d_k$ is the dimension of the key vectors.

This mechanism permits the model to allocate distinct significance to various words within the input sequence, aiding context-aware learning. The model then uses these learned weights to generate contextualized embeddings. Pre-training on vast text corpora helps in learning the optimal weights for this self-attention mechanism.

Subsequently, for fine-tuning, a task-specific loss function is applied to adapt the model to specific NLP tasks. This fine-tuning process can be described by \cite{kumar2023comparative}:

\begin{equation}
\mathcal{L}(\theta) = \mathcal{L}_{\text{task-specific}}(f_{\text{task-specific}}(h, \theta))
\end{equation}
Where \(\mathcal{L}(\theta)\) the loss function for the overall model with parameters \(\theta\), 
 \(\mathcal{L}_{\text{task-specific}}\) the loss function specific to the NLP task. 
 \(f_{\text{task-specific}}(h, \theta)\) the fine-tuned model for the task with parameters \(\theta\)
In this way, foundation models are not only mathematically profound but also highly adaptable for a wide spectrum of NLP tasks, making them transformative in the field of artificial intelligence.

\subsection{Utilizing Foundation Models for Food Security Applications}
\begin{figure}[htb]
\centering
\includegraphics[width=1\textwidth]{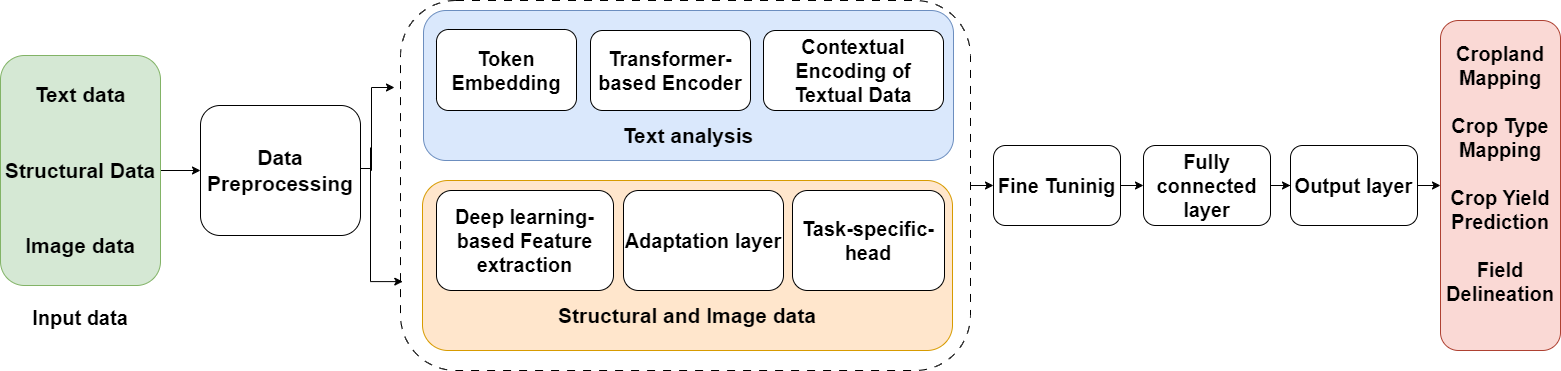} 
\caption{Block diagram of the proposed foundation model for Food Security applications.} 
\label{fbd}
\end{figure}

Food security is a pressing global concern, with far-reaching implications for human well-being and sustainable development. Land-use mapping, enabled by the integration of foundation models, has emerged as a vital tool in addressing food security challenges. Figure \ref{fbd} illustrates the block diagram of the proposed foundational model for food security applications. 

\subsection{Foundation Model Structures for Text Classification}
Foundation models for text classification are neural networks designed to process raw text data and produce classifications. The use of these models involves a blend of structural elements to attain cutting-edge outcomes in diverse NLP assignments.

\begin{enumerate}
\item \textbf{Input Layer}:
   The input to the model is raw text, which undergoes tokenization into subword or word-level tokens. Each token is represented as an embedding vector to create a dense, fixed-dimensional input representation.

\item \textbf{Embedding Layer}:
   The embedding layer transforms tokenized input into dense vectors, capturing semantic word meanings and contextual relationships. This layer employs advanced embedding techniques such as Word2Vec and subword embeddings.

\item \textbf{Transformer Architecture}:
   Foundation models rely on the transformer architecture, a fundamental structure for NLP. It comprises multiple transformer layers, typically stacked hierarchically. Each transformer layer comprises two sub-layers:

   \begin{enumerate}
   \item \textbf{Multi-Head Self-Attention Layer}:
      This sub-layer computes weighted representations of all words in the input text using self-attention mechanisms. For a given input sequence \(X\).
   \item \textbf{Position-wise Feedforward Layer}:
      Following self-attention, a position-wise feedforward neural network is applied to each position independently to capture local text patterns. For an input sequence \(X\), this can be represented as \cite{goodfellow2020generative}:
      \begin{equation}
    \text{FeedForward}(X) = \sigma(W_2\text{ReLU}(W_1X + b_1) + b_2)
      \end{equation}
      Where \(W_1\), \(W_2\), \(b_1\), and \(b_2\) are weight matrices and bias terms.
   \end{enumerate}

\item \textbf{Classification Head}:
   At the top of the model, a classification head is attached. It consists of fully connected layers that map transformer representations to class probabilities. The final output can be represented as \cite{goodfellow2016deep}:
   \begin{equation}
       P(Y|X) = \text{softmax}(WX + b)
  \end{equation}
   where \(Y\) is the class label, \(X\) is the transformer output, and \(W\) and \(b\) are weight and bias parameters.

\item \textbf{Training Objective}:
   Foundation models for text classification are trained using a cross-entropy loss function, quantifying the difference between predicted class probabilities and ground truth labels. The loss can be represented as \cite{goodfellow2016deep}:
   \begin{equation}
   \mathcal{L} = -\sum_{i} y_i \log(p_i)
  \end{equation}
   Where \(y_i\) is the ground truth label, \(p_i\) is the predicted probability, and the sum is taken over all classes.


\end{enumerate}
Foundation models, exemplified by BERT and its derivatives, have revolutionized NLP tasks by capitalizing on the transformer's self-attention mechanism. Their ability to effectively capture contextual information renders them powerful tools for a spectrum of text classification applications.
\subsection{Foundation Model Structures for Structured Data Classification}
Foundation models, celebrated for their versatility in various applications, extend their prowess to structured data classification. This elucidation delves into the intricacies of their architecture, unveiling the essential components and processes that render them particularly suitable for structured data.
\begin{enumerate}
\item \textbf{Input Data Representation}:
Structured data classification revolves around tabular data, where rows represent instances, and columns denote attributes. These attributes span a diverse array of types, including numerical, categorical, and textual data.

\item \textbf{Model Architecture}:

    \begin{enumerate}
    \item \textbf{Embedding Layer}:
    Foundation models initiate similarly to their NLP counterparts with an embedding layer. In this layer, categorical attributes are transformed into continuous vectors, while numerical features remain unaltered. Textual features can be embedded using techniques such as Word2Vec or pre-trained word embeddings.

    \item \textbf{Transformer Layers}:
    Foundation models leverage the transformer architecture, customized for structured data:

        \begin{enumerate}
        \item \textbf{Multi-Head Self-Attention}:
        Self-attention mechanisms identify relationships between rows and features within tabular data. The model autonomously learns the relevance of specific features and their dynamic interactions.

        \item \textbf{Position-wise Feedforward Networks}:
        Following self-attention, position-wise feedforward networks capture localized patterns and interactions within structured data.
        \end{enumerate}
    \end{enumerate}

\item \textbf{Classification Head}:
A classification head sits atop the model, orchestrating the prediction of class labels or target variables. This head typically incorporates fully connected layers, culminating in an output layer equipped with activation functions.

\item \textbf{Training and Objective}:
Foundation models for structured data classification undergo rigorous training, with the choice of a loss function tailored to the classification task. Binary classification often employs cross-entropy loss, while multi-class problems necessitate categorical cross-entropy. For regression tasks, mean squared error may be adopted as the loss function.


\end{enumerate}
These models offer numerous advantages, including the automatic learning of features, interpretability, and the potential for transfer learning across diverse domains. By harnessing the powerful capabilities of transformers, they unveil intricate feature interactions within tabular data, providing an innovative approach to solving classification challenges across various domains.
\subsection{Foundation Model Structures for Image Classification}
Foundation models have demonstrated exceptional performance in image classification tasks, equipped with a structured architecture optimized for the analysis of visual data. This explanation offers a systematic breakdown of the intricate components and scientific foundations governing the operation of foundation models for image classification.

\begin{enumerate}
\item \textbf{Input Data Representation}:
   Image classification focuses on visual data in the form of multi-dimensional arrays of pixel values, with channels representing color information.

\item \textbf{Model Architecture}:
   Foundation models for image classification consist of fundamental architectural components:
   \begin{enumerate}
   \item \textbf{Convolutional Neural Networks (CNNs)}:
      Central to image classification models, CNNs comprise layers like convolutional, pooling, and fully connected layers, which hierarchically learn features from input images.

   \item \textbf{Pre-trained Models}:
      Foundation models often integrate pre-trained architectures (e.g., VGG, ResNet, Inception) that have acquired rich image features from extensive datasets. These models serve as a foundational knowledge base and can be fine-tuned for specific classification tasks.

   \item \textbf{Fine-Tuning}:
      Fine-tuning is a critical process to adapt pre-trained models to image classification tasks. It customizes the model's learned weights to align with the specific dataset and classification objectives.

   \item \textbf{Transfer Learning}:
      Transfer learning is extensively used, enabling the transfer of knowledge learned from one task or dataset to another. Pre-trained model representations often serve as valuable features for image classification.

   \end{enumerate}

\item \textbf{Training and Objective}:
   Image classification models are trained through supervised learning. The choice of loss function depends on the number of classes and the nature of the problem. For multi-class classification, categorical cross-entropy is commonly used to quantify the disparity between predicted class probabilities and true class labels.

\item \textbf{Prediction and Activation Functions}:
   Prediction in image classification models is facilitated by activation functions, such as softmax, applied to the final layer's output. The softmax function converts raw scores into class probabilities. For an image \(I\) and class \(j\), the probability \(P(j|I)\) is computed using the softmax function.
\end{enumerate}
Foundation models for image classification epitomize the forefront of computer vision, providing a versatile and robust solution for tasks that span object recognition, medical image analysis, and beyond.
\begin{table}[h]
\centering
\caption{Comparison of Foundation Model Structures for Text, Structured Data, and Image Data}
\resizebox{\textwidth}{!}{
\renewcommand{\arraystretch}{1.2}
\begin{tabular}{|c|c|c|c|}
\hline
\textbf{Aspect} & \textbf{Text} & \textbf{Structured Data} & \textbf{Image Data} \\
\hline
\textbf{Input Data} & Raw text & Tabular data & Images \\
\hline
\textbf{Embedding Layer} & Embeds tokens & Converts categorical attributes & Utilizes pre-trained CNNs \\
\hline
\textbf{Transformer Layers} & Multi-head self-attention & Self-attention and feedforward & Pre-trained CNN layers \\
\hline
\textbf{Pre-training} & On extensive text corpora & Pre-training not common & On large image datasets \\
\hline
\textbf{Fine-Tuning} & Common for task-specific adaptation & Fine-tuning is typical & Fine-tuning for specific tasks \\
\hline
\textbf{Training Objective} & Cross-entropy loss & Variable loss functions & Cross-entropy or MSE \\
\hline
\textbf{Model Architecture} & Transformer-based & Utilizes pre-trained models & CNN-based \\
\hline
\textbf{Activation Function} & Softmax for classification & Depends on the task & Softmax for classification \\
\hline
\textbf{Data Types} & Textual data & Tabular data (numerical, categorical) & Image data (pixel values) \\
\hline
\end{tabular}
\label{f2}
}
\end{table}
Table \ref{f2} illustrates the distinctive characteristics of foundation model structures when applied to text, structured data, and image data. In the context of input data, text models process raw text, structured data models handle tabular data with numerical and categorical attributes, while image models work with pixel values from images. These models utilize different embedding techniques: text models embed tokens, structured data models transform categorical attributes, and image models rely on pre-trained CNNs. Furthermore, the training goals differ, with text models employing cross-entropy, structured data models using different loss functions, and image models utilizing cross-entropy for classification along with Mean Squared Error (MSE) for regression. The differences in model architecture, fine-tuning practices, and activation functions highlight the adaptability of foundation models to diverse data types.
\subsection{Enhancing Efficiency in Fine-Tuning}
Fine-tuning involves customizing pre-existing foundation models for specific NLP tasks, including sentiment analysis, text classification, and question-answering \cite{raffel2019exploring}. Without fine-tuning, foundation models might not achieve optimal performance on task-specific data, but the process can be resource-intensive due to the need to update a large number of model parameters.
Parameter efficiency is crucial in this context, as it involves achieving high task performance with a minimal increase in the number of model parameters. The goal of parameter-efficient fine-tuning is to make the most of as few model parameters as possible while maintaining high task performance. Several techniques support this approach:

Gradual unfreezing entails freezing most pre-trained model layers and unfreezing only a small subset, such as the last few layers or task-specific layers, reducing the number of parameters requiring adjustment. Knowledge distillation is the process of training a compact, specialized model to replicate the performance of a larger, pre-trained model, resulting in a reduction in the overall model size. Pruning techniques remove unimportant connections and parameters from the model using methods like magnitude-based pruning, preserving performance while decreasing parameter count. Architectural modifications, such as replacing self-attention layers with more efficient variants like sparse attention mechanisms, can optimize the foundation model's architecture for specific tasks. Quantization involves reducing the precision of model weights and activations, significantly decreasing model size and memory requirements while maintaining performance.

The benefits of parameter-efficient fine-tuning are substantial. By fine-tuning only a fraction of model parameters, you can execute models on less powerful hardware or deploy them on edge devices. Smaller models fine-tuned with parameter-efficient methods typically offer faster inference times, making them suitable for real-time applications. Furthermore, these techniques contribute to reducing the energy consumption and carbon footprint associated with operating large models in data centers, aligning with environmental and efficiency goals.

\section{Conclusions}
The application of foundation models marks a pivotal advancement in the sphere of food security, effectively tackling the intricate challenges of the world's food systems. The study has shed light on the adaptability of foundation models in diverse food security domains, spanning from crop type classification to cropland mapping, field delineation, and crop yield prediction. By leveraging various data types, such as remote sensing data, meteorological records, and historical databases, foundation models offer remarkable flexibility. They present a potent avenue to surmount the limitations associated with traditional deep learning and machine learning techniques, ensuring precise, granular insights that empower decision-makers.
Built upon extensive datasets and pre-trained on various domains, foundation models have ushered in a new era of food security applications. They bring enhanced accuracy, optimized resource allocation, and streamlined data processing. Furthermore, their capacity to adapt to a multitude of data sources and predictive tasks leads to the generation of invaluable insights, pivotal for the sustainable future of food production and distribution.
In the face of mounting global challenges in food security, encompassing factors like population growth and climate variations, foundation models are poised as a transformative catalyst. They equip policymakers, researchers, and farmers with the means to arrive at more informed decisions, manage resources effectively, and ensure the world's food supply. Through fostering interdisciplinary collaboration, embracing emerging technologies, and persistently refining these models, a path is forged toward a future in which food security is a reality, not a concern. This paper stands as a crucial milestone in realizing the potential of foundation models in the collective pursuit of global food security.


\bibliographystyle{elsarticle-num} 




\end{document}